\documentclass[10pt,letterpaper]{article}  
\usepackage{a4wide}
\usepackage{graphicx}
\usepackage{amssymb}
\usepackage{epstopdf}
\usepackage{graphics}
\usepackage{amsmath, amssymb}
\numberwithin{equation}{section}
\usepackage{graphicx}   
\usepackage{verbatim}   
\usepackage{color}      
\usepackage{subfig}  
\usepackage{hyperref}   
\usepackage{xspace}
\usepackage{float}
\usepackage{amsthm}

\DeclareMathOperator*{\argmax}{arg\,max}

\newcommand{\bff}{\mathbf{f}}

\newcommand{\bfy}{\mathbf{y}}

\newcommand{\bfx}{\mathbf{x}}

\newcommand{\bfI}{\mathbf{I}}
\newcommand{\bfW}{\mathbf{W}}
\newcommand{\bfw}{\mathbf{w}}

\newcommand{\bs}[1]{\boldsymbol #1}
\newcommand{\btheta}{\bs\theta}

\title{\large \bf
Manifold Relevance Determination: Learning the Latent Space of Robotics}
  
\author{Pete Trautman
}

\date{}

\begin{document}

\maketitle
\thispagestyle{empty}
\pagestyle{empty}

\begin{abstract}
\noindent In this article we present the basics of \emph{manifold relevance determination} (MRD) as introduced in~\cite{mrd}, and some applications where the technology might be of particular use.  Section~\ref{sec:intro-mrd} acts as a short tutorial of the ideas developed in~\cite{mrd}, while Section~\ref{sec:applications} presents possible applications in sensor fusion, multi-agent SLAM, and ``human-appropriate'' robot movement (e.g. legibility and predictability~\cite{dragan-hri-2013}).  In particular, we show how MRD can be used to construct the underlying models in a data driven manner, rather than directly leveraging first principles theories (e.g., physics, psychology) as is commonly the case for sensor fusion, SLAM, and human robot interaction.  We note that~\cite{mrd-grasping} leveraged MRD for correcting unstable robot grasps to stable robot grasps.
\end{abstract}

\section{What is MRD?}
\label{sec:intro-mrd}
\begin{figure}[htb]
      \centering
\includegraphics[scale=0.85]{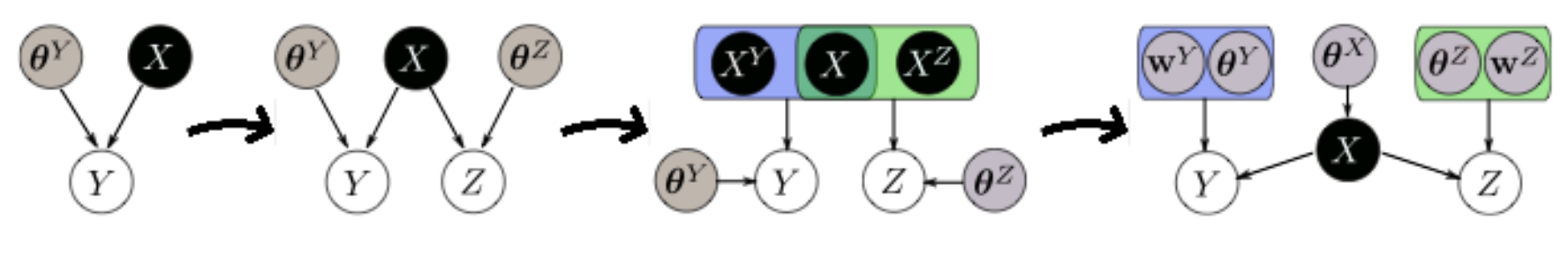}\vspace{-0mm}
      \caption{\small{\textbf{Illustration copied from~\cite{mrd}.}  Graphical model evolution of MRD, from left to right: first, we see GPLVM (alternatively, dual probabilistic PCA), which is then generalized to multiple views of the data over a single latent space.  In the third frame, we see the the introduction of shared and private latent spaces to explain variance due to a specific view.  Finally, MRD introduces an approximate latent space marginalization which in turn allows for \emph{automatic relevance determination} priors to be used, thus enabling soft boundaries between shared and private latent subspaces and data driven factorization of the shared subspace.} }
      \label{fig:mrd_representation}
\end{figure}

In this section, we explain how MRD has its origins in PCA---indeed, if we
\begin{itemize} 
\item generalize PCA to be nonlinear and probabilistic, we arrive at Gaussian process latent variable models (GPLVM).  
\item If GPLVMs are generalized to the case of multiple views of the data we arrive at shared GPLVMs.  
\item If we introduce private spaces to shared GPLVMs, then we recover a factorization of the latent space that encodes variance specific to the views.  
\item Finally, if we approximately marginalize the latent space (instead of optimizing the latent space) we can automatically (i.e., in a data driven manner) determine the dimensionality and factorization of the latent space.  
\end{itemize}
Automatic determination of the dimensionality and factorization of the nonlinear latent space from multiple views is \emph{manifold relevance determination}.  See Figure \ref{fig:mrd_representation} for the graphical model corresponding to this evolution.

\subsection{Generalizing PCA}
\label{sec:generalizing_pca}
To begin then, consider PCA: we are given a collection of centered observational data $Y = [\bfy_1, \ldots ,\bfy_N] \in \mathbb R^{N\times D}$, and we wish to relate this data to a latent space $X= [\bfx_1, \ldots ,\bfx_N]\in \mathbb R^{N\times q}$ (where hopefully $q \ll D$) via the linear embedding $\bfW\in \mathbb R^{D\times q}$
\begin{align}
\label{eq:linear_embed}
\bfy_n = \bfW\bfx_n.
\end{align}
For this formulation, the solution is both exact and efficient: compute the eigendecomposition of the covariance matrix $YY^{\top}$, and then construct $\bfW$ by choosing the $q$ eigenvectors with the largest eigenvalues.  PCA is thus interpreted as a linear projection of the data onto the subspace that most efficiently captures the variance in the data.  

Imagine now that we wish to formulate a probabilistic version of PCA; we can add a noise term to Equation \ref{eq:linear_embed}:
\begin{align}
\label{eq:ppca_embed}
\bfy_n = \bfW\bfx_n + \eta_n
\end{align}
where we assume
\begin{align}
p(\eta_n) = \mathcal N(\bf0, \beta^{-1}\bfI),
\end{align}
which allows us to formulate a likelihood over the data
\begin{align*}
p(\bfy_n  \mid \bfx_n, \bfW, \beta) = \mathcal N(\bfy_n  \mid \bfW\bfx_n , \beta^{-1}\bfI).
\end{align*}
At this stage, we have a choice to make---we can either
\begin{enumerate}
\item Marginalize the latent variables $X$ and optimize the parameters $\bfW$ or,
\item Marginalize the parameters $\bfW$ and optimize the latent variables $X$
\end{enumerate}
As it turns out, these two approaches are dual to one another (see \cite{gplvm}); the first approach, called probabilistic PCA (\cite{ppca}), reduces to PCA if one chooses the parameters $\bfW$ that maximize the marginal likelihood
\begin{align*}
\bfW^* = \argmax_{\bfW} p(Y \mid \bfW, \beta).
\end{align*}
Alternatively, if we marginalize the parameters (by choosing a Gaussian prior over the parameters, $p(\bfw) = \prod_i\mathcal N(\bfw_i\mid \bf0, \bfI)$), we arrive at a likelihood conditioned on the latent space:
\begin{align}
\label{eq:gplvm_marginal}
p(Y \mid X, \beta) &= \prod_{d=1}^Dp(\bfy_d \mid X, \beta) \nonumber\\
&= \prod_{d=1}^D \mathcal N(\bfy_d \mid \bf0, XX^{\top} + \beta^{-1}\bfI)
\end{align}
Optimization of the latent variables results in a matrix decomposition problem equivalent to finding the largest $q$ eigenvectors of $YY^{\top}$---once again, we recover PCA in its traditional form.

However, if instead we pause at Equation \ref{eq:gplvm_marginal} and recall the ``inner product'' kernel function of Gaussian processes (see \cite{gpmlras})
\begin{align}
\label{eq:gplvm_kernel}
k(\bfx_i, \bfx_j) = \bfx_i^{\top}\bfx_j + \beta^{-1}\delta_{ij},
\end{align}
and note that this kernel function $k(\bfx_i, \bfx_j)$ encodes \emph{linear embeddings of the data into the latent space}, we immediately see how the decomposition 
\[
\mathcal N(\bf{y}_d \mid \bf{0}, XX^{\top} + \beta^{-1}\bfI)
\] 
suggests a novel generalization of probabilistic PCA: by replacing the inner product kernel with a covariance function that allows for nonlinear functionality (thus nonlinear embedding functions), we recover a \emph{nonlinear, probabilistic} version of PCA.  This approach is called Gaussian process latent variable models.

\subsection{The rest of the story}
Building on the narrative of Section \ref{sec:generalizing_pca}, we now formulate the MRD model. 
\begin{itemize}
\item Suppose that we are given two ``views'' of a dataset, $Y \in \mathbb R^{N\times D_Y}$ and $Z \in \mathbb R^{N\times D_Z}$ (the number of samples $N$ in each view could be different for $Y$ and $Z$).
\item  We assume the existence of a single latent variable $X\in \mathbb R^{N\times q}$ that provides a low dimensional representation of the data through the nonlinear mappings
\begin{align}
\{ \bff_d^Y\}_{d=1}^{D_Y} \colon X \mapsto Y
\end{align}
and 
\begin{align*}
\{ \bff_d^Z\}_{d=1}^{D_Z} \colon X \mapsto Z.
\end{align*}
Further, the embeddings are corrupted by additive gaussian noise, so we have that
\begin{align*}
&y_{nd} = f_d^Y(\bfx_n) + \epsilon^Y_{nd}\\
&z_{nd} = f_d^Z(\bfx_n) + \epsilon^Z_{nd},
\end{align*}
where $\epsilon^{Y,Z } \sim \mathcal N(\bf0, \sigma^{\{Y,Z \}})$, and $y_{nd},z_{nd}$ represents dimension $d$ of point $n$.
\item Similar to Equations~\ref{eq:ppca_embed} and~\ref{eq:gplvm_marginal}, the above bullets lead to a likelihood function of the form
\begin{align}
\label{eq:mrd_likelihood}
p(Y,Z \mid X, \btheta )
\end{align}
where $\btheta = \{\btheta^Y, \btheta^Z \}$ denotes the parameters of the mapping functions $\bff_d^Y$ and $\bff_d^Z$ and the noise variances $\epsilon^{\{Y,Z \}}$.
\end{itemize}
Just as in Section \ref{sec:generalizing_pca}, we are forced to make a choice at this point---that is, we need to compute the likelihood \ref{eq:mrd_likelihood}, but we cannot marginalize over both the latent space and parameters.

Thus, we make some modeling choices:
\begin{itemize}
\item As with GPLVMs, we place a Gaussian process prior over the embedding mappings $\bff^Y_d$ and $\bff^Z_d$ ($K^Y$ is as introduced in Equation \ref{eq:gplvm_kernel}, but evaluated on $X$):
\begin{align}
\label{eq:sensor_models}
p(\{ \bff^Y_d\}_{d=1}^{D_Y} \mid X, \btheta^Y) = \prod_{d=1}^{D_Y}\mathcal N(\bff^Y_d \mid \bf0, K^Y),
\end{align}
and similarly for $\{ \bff^Z_d\}_{d=1}^{D_Z}$.  This allows us to model the embedding mappings \emph{non parametrically}, and, further, allows us to analytically marginalize out the parameters (or mappings):
\begin{align}
p(Y,Z \mid X, \btheta ) = \int p(Y \mid \{ \bff^Y_d\}) p(\{\bff^Y_d\}\mid X, \btheta^Y)
p(Z \mid  \{ \bff^Z_d\}) p(\{\bff^Z_d\}\mid X, \btheta^Z)
\end{align}
where $\{ \bff^Y_d\},\{ \bff^Z_d\}$ is shorthand for $\{ \bff^Y_d\}_{d=1}^{D_Y},\{ \bff^Z_d\}_{d=1}^{D_Z}$.
\item Unfortunately, while we can marginalize the mappings using GPs, we cannot analytically marginalize the latent space; this is a critical point---\emph{not marginalizing the latent space forces us to choose the dimensionality of the latent space, and forces us to choose how the latent space factorizes over shared and private subspaces.}
\item One of the key contributions of MRD is \emph{approximately marginalizing} the latent space using variational methods.  This enables the use of ``automatic relevance determination'' (ARD) GP prior kernels; ARD kernels, in turn, enable the dimensionality and the factorization of the latent space to be determined from the data. 
\end{itemize}
\subsection{Advantages of MRD}
\begin{enumerate}
\item The method models nonlinear embeddings into a factorized latent space.  The prior over the embeddings is a GP, which is very expressive, and can thus capture a wide variety of embedding functions.  The dimensionality of the latent space is driven by the data (rather than a heuristic).  The factorization of the latent space is also driven by the data (rather than a heuristically imposed boundary on the latent space).  
\item It is an unsupervised approach.  For the purposes of multi agent SLAM, one could imagine implementing it as a ``featureless'' approach to map building.  The latent space would conceivably correspond to the 6 dof pose of the vehicle, similar to the interpretation presented in the Yale Faces Experiment of \cite{mrd}.
\item The method is fully probabilistic.  Thus, when one tries to regress across the latent space (perhaps in between views or apart from available training data), the answer is a distribution, rather than just a match or not a match.
\item Similarly, when one projects back out into viewpoint space, the answer is again a distribution---so the question ``how does this novel view correspond to the training views'' is answered with a distribution.
\end{enumerate}


%

\section{Applications of MRD}
\label{sec:applications}
Inspired by the work in~\cite{mrd-grasping}, which applied MRD to the problem of transferring between stable and unstable robot grasps, we suggest the following potential applications.
\subsection{MRD for Sensor Fusion}
MRD naturally captures the concept of sensor fusion; imagine that one has $\ell$ views of some target $\{Y^1, \ldots, Y^{\ell}\}$.   The traditional sequential Bayesian formulation of sensor fusion requires that we know a number of things in advance:
\begin{enumerate}
\item The state of the target.
\item The $\ell$ different sensor models.
\item What part of the state each sensor model captures.
\item \emph{The hard part}: how each sensor interacts with each other sensor during observation.
\end{enumerate} 
Notably, MRD learns all of these things directly from the $\ell$ views of the target:
\begin{enumerate}
\item The latent space $X$ that is learned during training corresponds to the target under observation.
\item The sensor models are learned using Gaussian process regression, as in Equation~\ref{eq:sensor_models}.
\item The private subspace $X_p^i$ captures what part of the target is being uniquely observed by sensor $Y^i$.
\item \emph{The hard part}: the subspaces shared between views $i$ and $j$ $X_s^{ij}$ correspond to when sensor modalities are being fused.  See the discussion of the Yale Face Experiment in \cite{mrd}.
\end{enumerate}
\subsection{MRD for Multi-Agent SLAM}
\begin{figure}[h]
      \centering
\includegraphics[scale=0.6]{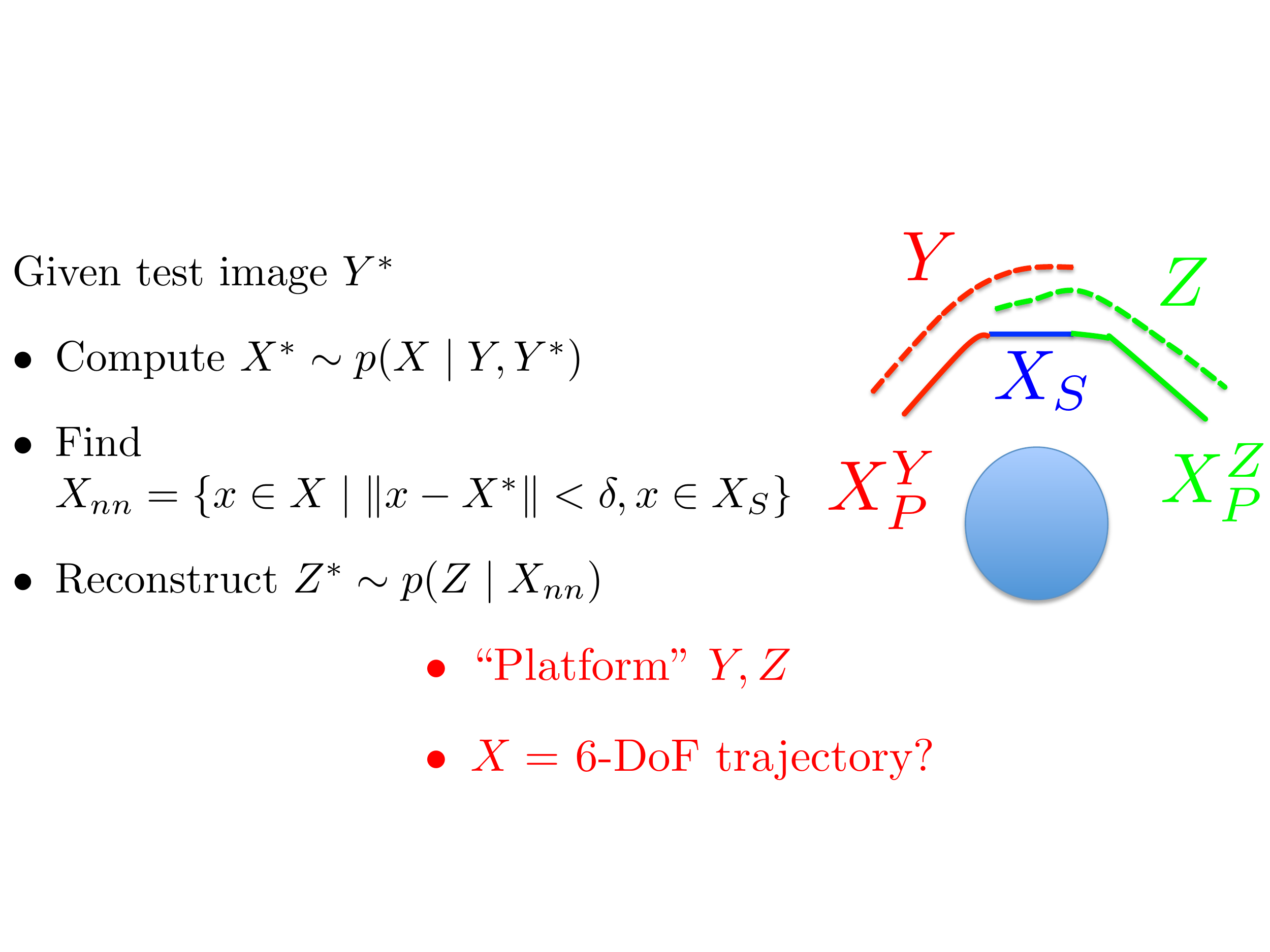}\vspace{-0mm}
      \caption{\small{Two agents ($Y,Z$) performing multi-agent SLAM via MRD.   The blue circle represents the scene of interest; the dashed colored lines represent the observations by either platform (red is agent $Y$, green is agent $Z$).  The latent space is represented by $X$, which is partitioned into two ``private'' subspaces ($X^Y_P$ and $X^Z_P$) and a ``shared'' latent subspace $X_S$.  The shared portion of the latent space allows us to transfer between the two observations $Y$ and $Z$.}}
      \label{fig:mrd_ma_slam}
\end{figure}

Similar to sensor fusion, multi-agent SLAM is a natural application of MRD.  In particular, imagine that we have two sensors $Y$ and $Z$ observing a scene (see Figure~\ref{fig:mrd_ma_slam}).  The two agents collect their datasets of images (let's use images for convenience's sake), and MRD is run.  Presumably, the latent space should correspond to the 6DOF pose of either of the cameras---in other words, the localization of the sensor.  

Now imagine that a third sensor enters the scene, and observes $Y^*$.  A sensible question in the context of MA-SLAM is ``what was the trajectory of the third sensor''?  This question is answered in the following way:
\begin{itemize}
\item We construct the distribution over possible poses of the camera $X^* \sim p(X \mid Y, Y^*)$.  In particular, we can query the distribution for the most likely pose of the new images $Y^*$:
\begin{align*}
X^{max} = \argmax_X p(X \mid Y, Y^*).
\end{align*}
\item $X^{max}$ corresponds to the most likely trajectory that generated the new set of sensor images $Y^*$.  Perhaps more importantly, $p(X \mid Y, Y^*)$ is actually the \emph{tracking density} of the third sensor!
\end{itemize}
We could additionally answer the following interesting question: given the new view $Y^*$, what would this portion of the scene look like to platform $Z$?  In other words, given the view $Y^*$, we can reconstruct what the scene would have looked like from sensor $Z$'s perspective.
\begin{itemize}
\item Start with $p(X \mid Y, Y^*)$.
\item  We next search the latent space for the nearest neighbors of $X^*$:
\begin{align*}
X_{nn} = \{ x\in X \mid \| x-X^*\| < \delta, x \in X_S\}
\end{align*}
\item We finally reconstruct the distribution $Z^* \sim p(Z \mid X_{nn})$; this distribution encodes what the portion of the scene observed by $Y^*$ would look like to sensor $Z$.
\end{itemize}
\subsection{MRD for ``human friendly'' movement}
We finish by discussing a trajectory generation problem.  The setting could be the following: imagine a robot arm trying to work collaboratively with a human in a grasping scenario. Perhaps it is as simple as tasking the robot and the human with simultaneous grasping of two objects (human friendly trajectories would be those that are both Legible and Predictable to the human, for instance).  

How do we accomplish this using MRD?  The idea is the following:
\subsubsection{Regressing over the latent space}  
\begin{figure}[h]
      \centering
\includegraphics[scale=0.5]{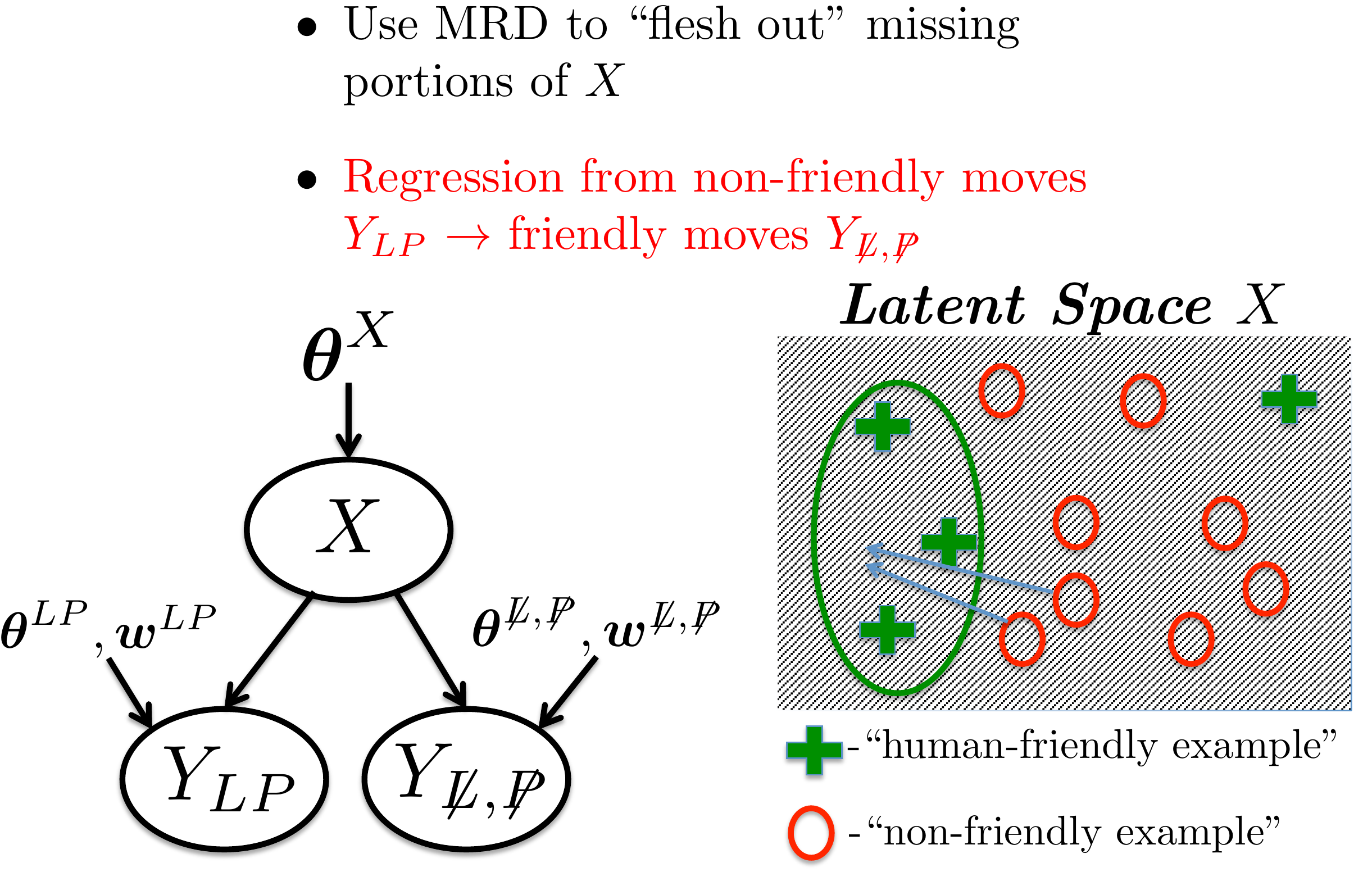}\vspace{-0mm}
      \caption{Regressing across small sample sets using MRD.}
      \label{fig:mrd_regress}
\end{figure}

\begin{itemize}
\item Collect samples $Y_{LP}$ and $Y_{\not L, \not P}$ that correspond, respectively, to human friendly and not human friendly trajectories. 
\item Use the regression capability of the underlying GPs to ``flesh out'' the latent space (see Figure~\ref{fig:mrd_regress}).  This is valuable because collecting examples of human friendly trajectories can be very costly---it might even involve a human demonstrator.  The underlying GP formulation provides a robust regression capability that can optimally utilize the data provided.  Furthermore, the probabilistic formulation informs the user when the system is uncertain about whether a new sample is human friendly or not.
\end{itemize}
\subsubsection{Transfer from random trajectory to desired trajectory}
Once we have collected the training samples, we hope to generate human friendly trajectories in an efficient and cheap manner.  We suggest the following approach (see Figure~\ref{fig:mrd_transfer}):
\begin{itemize}
\item use a COTs trajectory planner (RRT, PRM, CHOMP, etc) to generate a ``human ignorant'' path $\bar\bff^R$
\item Do the following:
\subitem Compute $X^* \sim p(X \mid Y_{\not L, \not P}, \bar\bff^R)$.
\subitem Compute the nearest neighbor set $X_{nn} = \{\tilde x\in X_s \mid \|\tilde x - X^* \| < \delta\}$
\subitem Compute the ``human friendly distribution'' $p(Y_{LP} \mid X_{nn})$.
\end{itemize}
Now, we choose 
\[
\bff^R_{max} = \argmax_{Y_{LP}} p(Y_{LP} \mid X_{nn})
\]
as our human friendly trajectory.  

The benefit of this approach is that it 1) leverages the ability of MRD to regress across small training sets (i.e. when samples are expensive) and also to 2) \emph{transfer} between generalized modes of operation, such as human-friendly and not-human-friendly.  One can also imagine applying this method to things like safe and not-safe modes.

\bibliographystyle{apalike}
{\footnotesize
\bibliography{mrd-bib}

\begin{thebibliography}{}

\bibitem[Bekiroglu et~al., 2016]{mrd-grasping}
Bekiroglu, Y., Damianou, A., Detry, R., Stork, J., Kragic, D., and Ek, C.
  (2016).
\newblock Probabilistic consolidation of grasp experience.
\newblock In {\em ICRA}.

\bibitem[Damianou et~al., 2012]{mrd}
Damianou, A., Ek, C., Titsias, M., and Lawrence, N. (2012).
\newblock Manifold relevance determination.
\newblock In {\em Proceedings of the 29th International Conference on Machine
  Learning}.

\bibitem[Dragan et~al., 2013]{dragan-hri-2013}
Dragan, A., Lee, K., and Srinivasa, S. (2013).
\newblock Legibility and predictability of robot motion.
\newblock In {\em International Conference on Human-Robot Interaction}.

\bibitem[Lawrence, 2005]{gplvm}
Lawrence, N. (2005).
\newblock Probabilistic non-linear principal component analysis with gaussian
  process latent variable models.
\newblock {\em The Journal of Machine Learning Research}.

\bibitem[Rasmussen and Williams, 2006]{gpmlras}
Rasmussen, C.~E. and Williams, C. (2006).
\newblock {\em Gaussian Processes for Machine Learning}.
\newblock MIT Press.

\bibitem[Tipping and Bishop, 1999]{ppca}
Tipping, M.~E. and Bishop, C.~M. (1999).
\newblock Probabilistic principal component analysis.
\newblock {\em Journal of the Royal Statistical Society, Series B}.

\end{thebibliography}
}

\end{document}